\pdfoutput=1

\documentclass[11pt]{article}

\usepackage{ACL2023}

\usepackage{times}
\usepackage{latexsym}

\usepackage[T1]{fontenc}

\usepackage[utf8]{inputenc}

\usepackage{times}
\usepackage{latexsym}
\usepackage{algorithm, algorithmic}
\usepackage{amsmath} 
\usepackage{amssymb}
\usepackage{multirow}
\usepackage{arydshln} 
\usepackage{makecell}
\usepackage{hhline}
\usepackage{xspace}
\usepackage{booktabs}
\usepackage{float}
\usepackage{graphicx}

\usepackage{amssymb}
\usepackage{pifont}
\usepackage{bm}
\usepackage{enumitem}
\usepackage{diagbox}
\setlist[itemize]{leftmargin=*}
\setitemize[1]{itemsep=0pt,partopsep=0pt,parsep=\parskip,topsep=0pt}
\usepackage{microtype}

\usepackage{soul} 

\usepackage{color, xcolor} %

\usepackage{inconsolata}

\title{Is ChatGPT a Highly Fluent Grammatical Error Correction System? A Comprehensive Evaluation}

\author{
        Tao Fang$^1$~~~
       Shu Yang$^1$~~~ 
       Kaixin Lan$^1$~~~
        Derek F. Wong$^1\footnotemark[1]$\thanks{~~Corresponding Author}~~~
        Jinpeng Hu$^2$~~~ \\
        \textbf{Lidia S. Chao$^1$}~~~
        \textbf{Yue Zhang$^3$}~~~  \\
    $^1$NLP$^2$CT Lab, Department of Computer and Information Science, University of Macau \\
      \texttt{nlp2ct.\{taofang,shuyang,kaixin\}@gmail.com, \{derekfw,lidiasc\}@um.edu.mo} \\
      $^2$The Chinese University of Hong Kong (Shenzhen)~~~~$^3$Soochow University \\
      \texttt{jinpenghu@link.cuhk.edu.cn}~~~~
      \texttt{yzhang21@stu.suda.edu.cn}
    }

\begin{document}
\maketitle
\begin{abstract}

ChatGPT, a large-scale language model based on the advanced GPT-3.5 architecture, has shown remarkable potential in various Natural Language Processing (NLP) tasks. However, there is currently a dearth of comprehensive study exploring its potential in the area of Grammatical Error Correction (GEC). To showcase its capabilities in GEC, we design zero-shot chain-of-thought (CoT) and few-shot CoT settings using in-context learning for ChatGPT. Our evaluation involves assessing ChatGPT's performance on five official test sets in three different languages, along with three document-level GEC test sets in English. Our experimental results and human evaluations demonstrate that ChatGPT has excellent error detection capabilities and can freely correct errors to make the corrected sentences very fluent, possibly due to its over-correction tendencies and not adhering to the principle of minimal edits. Additionally, its performance in non-English and low-resource settings highlights its potential in multilingual GEC tasks. 
However, further analysis of various types of errors at the document-level has shown that ChatGPT cannot effectively correct agreement, coreference, tense errors across sentences, and cross-sentence boundary errors.

\end{abstract}

\section{Introduction}
In recent years, Natural Language Processing (NLP) has made significant advancements due to the emergence of large language models (LLMs). 
Among the numerous LLMs available, Generative Per-trained Transformer (GPT) models \citep{radford2019language, NEURIPS2020_1457c0d6} have demonstrated high efficacy in various NLP tasks.
Recently, ChatGPT\footnote{\url{https://openai.com/blog/chatgpt}}, an advanced language model developed by OpenAI, has gained significant attention from researchers and practitioners in the field of NLP \citep{qin2023chatgpt,luo2023chatgpt,liu2023comprehensive,hendy2023good}. 
Building upon the InstructGPT \citep{ouyang2022training}, ChatGPT is a groundbreaking innovation in the field of conversational agents.
It possesses the remarkable ability to understand complex instructions and generate responses that closely mimic human speech. In addition to its conversational abilities, ChatGPT has demonstrated impressive performance in various other NLP tasks, including machine translation \cite{jiao2023chatgpt,hendy2023good, Peng2023ChatGPT4MT},question-answering \citep{bang2023multitask}, and text summarization \citep{yang2023exploring}. 

When it comes to the Grammatical Error Correction (GEC) ability of ChatGPT, it appears that many individuals favor its use for text revision and refinement. 
Nonetheless, there is currently a scarcity of comprehensive research literature regarding ChatGPT's genuine error correction capabilities. 
To the best of our knowledge, only one study \citep{wu2023chatgpt} conducts a preliminary evaluation of ChatGPT's performance in the English GEC task of the CoNLL14 test by only analyzing random 300 sentences and yields some preliminary but not sufficient results, which leaves us unclear about ChatGPT's specific capabilities and advantages in the GEC task. To address the current research gap, the aim of this study is to comprehensively investigate the performance and potential of ChatGPT in GEC, as well as to compare it with state-of-the-art (SOTA) models. This includes evaluating its performance at both the sentence-level and document-level GEC, designing prompts for zero-shot and few-shot scenarios, and assessing its effectiveness in English, non-English languages, and low-resource settings.

To explore the potential of ChaGPT for GEC, we first conduct a preliminary study to examine the effectiveness of our designed prompting methods in both \textbf{zero-shot} and \textbf{zero-shot chain-of-thought (CoT)} settings \citep{kojima2023large}. Our findings indicate that incorporating CoT techniques can significantly enhance the performance of GEC on both CoNLL14 test and BEA19 test sets. Moreover, we propose the utilization of few-shot prompts in conjunction with CoT techniques to further enhance ChatGPT's performance through in-context learning \cite{NEURIPS2020_1457c0d6}. We evaluate ChatGPT's performance for sentence-level grammatical error correction (GEC) in three languages covered five official test sets using \textbf{zero-shot CoT}, \textbf{1-shot CoT}, \textbf{3-shot CoT}, and \textbf{5-shot CoT}. Besides, we also assess its performance in \textbf{zero-shot CoT} and \textbf{1-shot CoT} scenarios for document-level GEC.

Furthermore, we conducte both automatic human evaluation and manual human evaluations, comparing ChatGPT to mainstream SOTA models (GECToR and T5) and a widely-used commercial GEC system, Grammarly. These assessments provide valuable and reliable insights into the strengths and weaknesses of ChatGPT for the GEC task. We also conducte an analysis of error types in the document-level GEC, revealing the reasons behind ChatGPT's suboptimal performance in this context.

Based on the results of our experiments and analyses, we summarize the following obervations: \\

\begin{itemize}
  \item ChatGPT exhibits a significant disparity with the current SOTA systems in terms of Precision and F$_{0.5}$, while its Recall performance is remarkably superior. This suggests that ChatGPT has the ability to detect errors in the text, but its elevated degree of modification freedom may result in superfluous changes. \\
    \item ChatGPT shows significant potential for GEC through zero-shot CoT and few-shot CoT strategies. Even on the JFLEG test set, it has shown minimal difference from the SOTA models and surpassed human evaluations, exhibiting human-like fluency. \\
    \item ChatGPT also demonstrates advantages in non-English languages and low-resource environments under the zero-shot CoT strategy, outperforming Transformer models trained from scratch. This highlights the potential of ChatGPT for multilingual GEC tasks. \\
    \item Our human evaluations indicate that ChatGPT performs with greater fluency in correcting grammatical errors, albeit with a noticeable tendency towards over-correction. Additionally, as sentence length increases, ChatGPT exhibits a tendency to follow minimal edits and the under-correction rate gradually increases. \\
     \item Upon analyzing the error types for document-level GEC, ChatGPT shows relatively poor performance in correcting agreement, coreference, and tense errors across sentences, as well as cross-sentence boundary errors. This could be due to ChatGPT's inherent limitations in processing excessively long sentences.
\end{itemize}

\section{Experimental Setup}

\subsection{Dataset}

\begin{table}[!t]
\small
\centering
\resizebox{.49\textwidth}{!}{
\begin{tabular}{l|l|l|l|l}
\toprule
\textbf{Lan.} & \textbf{Data} & \textbf{\#Sents}& \textbf{\#Docs} & \textbf{\#Doc.len.} \\
\cmidrule(lr){1-5}
\multirow{4}{*}{\textsc{EN}}
& CoNLL14 &  1,312 & 50 & 26 \\
& JFLEG test & 747  & - & - \\
& BEA19 test & 4,477 & - & - \\
& BEA19 dev & -  & 350 & 13 \\
& FCE test & -  & 194 & 14 \\
\cmidrule(lr){1-5}
\textsc{DE}
& Falko-MERLIN & 2,337 & - & - \\
\cmidrule(lr){1-5}
\textsc{ZH}
& NLPCC18 test & 2,000 & - & - \\
\bottomrule
\end{tabular}}
\linespread{1}
\caption{The datasets used in the evaluation. 
}
\label{Tab:datasets}
\end{table}

Our evaluation involves conducting sentence-level grammatical error correction (GEC) evaluations on a total of five official test sets across three languages: English, German, and Chinese. For English, we select the wildly-used CoNLL14 \citep{ng-etal-2014-conll} and BEA19 \citep{bryant-etal-2019-bea} GEC test sets. We use the official test set of NLPCC18~\cite{GEC_NLPCC} and the official Falko-MERLIN \citep{boyd-etal-2014-merlin} test set for Chinese and German, respectively. The four aforementioned test sets only contain minimal edits that rectify the grammatical errors in a sentence, without necessarily improving fluency or naturalness of the sentence. To assess the error correction abilities of GEC systems more accurately, we also evaluate on the JFLEG \citep{napoles-etal-2017-jfleg} test set in English, which represents a range of language proficiency levels and utilizes comprehensive fluency edits to enhance the accuracy of the evaluation.
Additionally, we contemplate conducting evaluations on three document-level test sets for the English language, following the methodology suggested by \citet{yuan-bryant-2021-document}. These test sets include the FCE document-level test set \citep{yannakoudakis-etal-2011-new}, the BEA19 document-level development set (\citep{bryant-etal-2019-bea}), and the CoNLL14 document-level test set \citep{ng-etal-2014-conll}. Table~\ref{Tab:datasets} presents the statistics of the datasets we used.

\subsection{Grammatical Error Correction Systems}
Currently, among the publicly available GEC systems for sentence-level GEC tasks, the state-of-the-art (SOTA) Seq2Seq GEC model is (m)T5 and its variant models \citep{rothe-etal-2021-simple}, and the SOTA Seq2Edit GEC model is the GECToR model \citep{omelianchuk-etal-2020-gector}. 
In addition to these SOTA models, we include the Transformer-base \citep{vaswani2017attention} GEC model as our baseline model for comparing the performance of the ChatGPT system. 
Since TagGEC \citep{stahlberg-kumar-2021-synthetic} is the SOTA model on the JFLEG test set, we also include it in the comparison. 
Regarding the comparison of the document-level GEC system, we utilize the MultiEnc-dec model proposed by \citet{yuan-bryant-2021-document}, which is presently considered as the SOTA for English document-level GEC task.

\subsection{ChatGPT system}
OpenAI has recently developed several GPT-3.5 series models\footnote{\url{https://platform.openai.com/docs/model-index-for-researchers}}, among which ChatGPT (gpt-3.5-turbo) stands out as the most advanced and specifically optimized for chat functionality. We assess the ChatGPT's performance in the GEC task using official API\footnote{\url{https://platform.openai.com/docs/api-reference}}. 
  
\subsection{Evaluation Method}
\paragraph{Sentence-Level Evaluation}
GEC systems are evaluated using automated metrics, which compare their output against gold-standard corrections from reference corpora. The selection of automatic evaluation metrics depends on their correlation with human judgments on different types of test sets. Therefore, following previous work \citep{katsumata-komachi-2020-stronger, omelianchuk-etal-2020-gector, rothe-etal-2021-simple}, we utilize the M2 Scorer\footnote{\url{https://github.com/nusnlp/m2scorer}} \citep{dahlmeier-ng-2012-better} to evaluate the performance of systems on CoNLL14 English, Falko-MERLIN German, and NLPCC18 Chinese GEC tasks. For assessing the BEA2019 test set, we employ an official ERRANT Scorer\footnote{\url{https://github.com/chrisjbryant/errant}} \citep{bryant-etal-2019-bea}. Additionally, we adopt the GLUE metric\footnote{\url{https://github.com/cnap/gec-ranking/}} \citep{napoles2016gleu} to evaluate the JFLEG test set.

\paragraph{Document-Level Evaluation}
Based on the approach by \citet{yuan-bryant-2021-document}, we manually process the raw data to produce document-level references since the available references are only at the sentence level. To evaluate the performance of the ChatGPT system on document-level GEC task, we adopt their method and utilize the official scorer of the BEA19 shared task \citep{bryant-etal-2019-bea}, the ERRANT Scorer \citep{bryant-etal-2017-automatic}, for document-level GEC evaluation. For further details, it can be referred to their repository\footnote{\url{https://github.com/chrisjbryant/doc-gec}}.

\paragraph{Human Evaluation}
Automatic evaluation may introduce bias in reflecting the true performance of a system, especially when there is a limited number of available references. Therefore, we conducte two types of human evaluations, namely automatic human evaluation and manual human evaluation, to more accurately assess the performance of the systems. 
To conduct automatic human evaluation, we follow the method of \citet{bryant-ng-2015-far} and \citet{napoles-etal-2017-jfleg} to compare the performance of a GEC system against human performance. They measure human performance on the CoNLL14 and JFLEG test sets, respectively, using 10 and 4 sets of human annotations. Furthermore, to further analyze the potential of the ChatGPT system, we invite three annotators to manually annotate the outputs of SOTA GEC systems from various aspects on the CoNLL14 test set. The detailed results are provided in Section~\ref{human}.

\begin{table}[ht]
\small
\centering
\resizebox{0.46\textwidth}{!}{
\begin{tabular}{p{1.9cm} | p{6cm}}
\toprule
 \multicolumn{1}{c|}{\textbf{Method}} &  \multicolumn{1}{c}{\textbf{GEC Prompt}} \\
\cmidrule(lr){1-2}
\multirow{10}{*}{\textbf{Zero-shot}} &"role": "system", "content": 'You are an English grammatical error correction tool that can identify and correct grammatical errors in a text.' \\&
 \texttt{"role": "user", "content": 'Please identify and correct any grammatical errors in the following sentence while keeping the original sentence structure unchanged as much as possible: '}\\ 
\cmidrule(lr){1-2}
\multirow{21}{*}{\textbf{Zero-shot CoT}} &"role": "system", "content": 'You are an English grammatical error correction tool that can identify and correct grammatical errors in a text.' \\&
\texttt{"role": "user", "content": 'Please identify and correct any grammatical errors in the following sentence indicated by <input> ERROR </input> tag, you need to comprehend the sentence as a whole before identifying and correcting any errors step by step while keeping the original sentence structure unchanged as much as possible. Afterward, output the corrected version directly without any explanations. Remember to format your corrected output results with the tag <output> Your Corrected Version </output>. Please start: <input> ERROR </input>: ' }\\
\bottomrule
\end{tabular}}
\linespread{1}
\caption{Zero-shot and zero-shot CoT settings for ChatGPT.}
\label{Tab:GEC_zero_shot_prompts}
\end{table}
\begin{table}[t]
\small
\centering
\resizebox{0.45\textwidth}{!}{
\begin{tabular}{l|ccc|ccc}
\toprule
{\multirow{2}{*}{\textbf{\makecell[l]{\\Method}}}}
& \multicolumn{3}{c|}{\textbf{CoNLL14}}
& \multicolumn{3}{c}{\textbf{BEA19 test}} \\
\cmidrule(r){2-7}
 & {Pre.}  & {Rec.}  & \textsc{F$_{0.5}$}   & {Pre.}  & {Rec.}  & \textsc{F$_{0.5}$}  \\

\cmidrule(lr){1-7}

Zero-shot &48.5 &58.9 &50.3   & 30.5 & 69.0 & 34.4 \\
~~~+ CoT & \bf 50.2 & \bf 59.0 & \bf 51.7$^{\dagger}$    &\bf 32.2 & \bf 70.5 & \bf 36.1 \\

\bottomrule
\end{tabular}} 
\linespread{1}
\caption{Performance of zero-shot for ChatGPT with CoT
prompting methods on the CoNLL14 and BEA19 English test sets. \textbf{Bold} values indicate the best Pre/ Rec/ F$_{0.5}$ scores. Statistically significant improvements over zero-shot prompting method, as indicated by $P${\_}value, $^{\dagger}p<0.01$. Note that the BEA19 test set is a blind test.
}
\label{Tab:GEC_zero_shot}
\end{table}

\section{Experiments}

\subsection{The Effect of Zero-Shot Prompts for ChatGPT}
\paragraph{Zero-shot Prompts Design} We explore the zero-shot capabilities of ChatGPT in GEC from two aspects: zero-shot and zero-shot CoT settings. Table~\ref{Tab:GEC_zero_shot_prompts} presents the different designed prompting methods.
Specifically, for zero-shot, we directly ask the ChatGPT system to \texttt{identify and correct any grammatical errors in the sentence while keeping the original sentence structure unchanged as much as possible.} 
However, we observe that using this normal prompts, ChatGPT tends to generate a multitude of explanations and the format of the resulting answers is often disorganized, necessitating manual intervention. 
To alleviate this issue, we adopt the zero-shot CoT technique \citep{kojima2022large} and design our own zero-shot CoT prompt for ChatGPT. 
Different from the zero-shot prompt, we use special tag \texttt{<input> Input Sentence </input>} to indicate the input sentences, and tell ChatGPT should output the results using the format \texttt{<output> Your Corrected Version </output>}. 
We instruct ChatGPT to approach the task by comprehending the sentence as a whole before identifying and correcting any errors step by step. Afterward, we request ChatGPT to provide us with the corrected sentences directly, \texttt{without any explanations}. We show some examples of using zero-shot and zero-shot CoT settings for ChatGPT in Appendix~\ref{appendix:zero-shot-settings}.

\paragraph{Preliminary Experimental Results}
We conduct a preliminary experiment in the CoNLL14 and BEA19 English test sets for ChatGPT to evaluate our designed zero-shot prompting methods, with the results reported in Table \ref{Tab:GEC_zero_shot}. 
We can observe that the \textbf{zero-shot + CoT} designed method significantly outperforms the normal \textbf{zero-shot} method both on the CoNLL14 and BEA19 test sets. 
Therefore, the subsequent zero-shot and few-shot experiments in all languages will be combined with the \textbf{CoT} prompting method.

\subsection{Few-Shot Prompts Design for ChatGPT}
\begin{figure*}[ht]
\centering
\includegraphics[width=0.96\textwidth, trim=0 0 0 0]{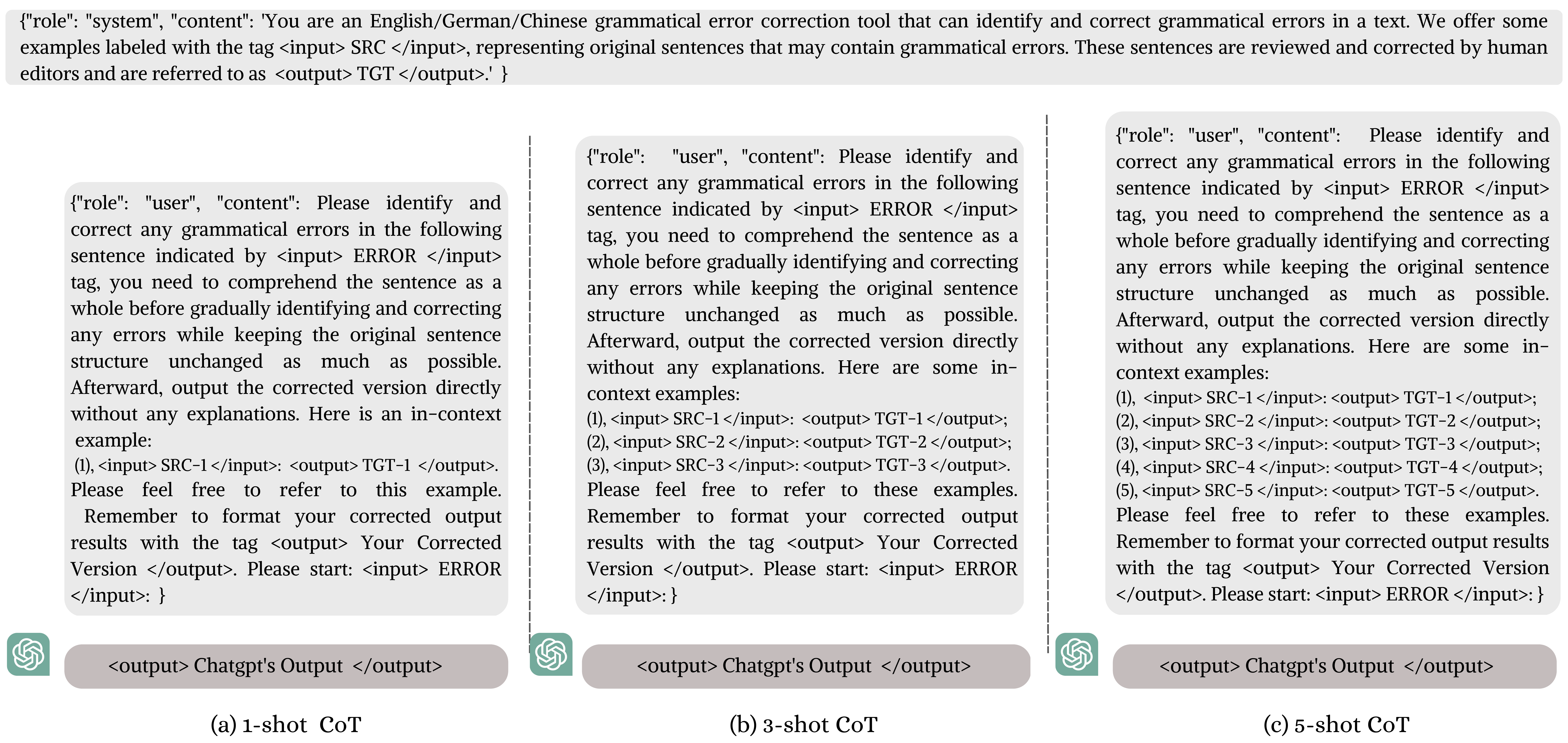}
\caption{An illustration of all few-shot CoT prompts used for ChatGPT to perform Grammatical Error Correction.}
\label{fig:few-shot}
\end{figure*}

Several studies have shown that including few labeled examples in the test input can enhance the performance of LLMs through in-context learning \cite{NEURIPS2020_1457c0d6}. 
In this section, we follow \citet{vilar2022prompting} to randomly select different in-context labeled examples from development data, as they indicate that different few-shot selection strategies are not necessarily superior to random selection. 
In particular, we randomly select input-output samples from the CoNLL13 test, BEA19 development, JFLEG development, Falko-MERLIN development, and NLCC18 train sets for the CoNLL14 test, BEA-19 test, JFLEG test, Falko-MERLIN test, and NLPCC18 test sets, respectively. 
Figure~\ref{fig:few-shot} shows the designed GEC few-shot prompts combined with the CoT technique for ChatGPT. 
In all cases, we test the performance with 1-shot CoT, 3-shot CoT, and 5-shot CoT for all languages.

\subsection{ChatGPT Performance on English GEC Task}
\begin{table*}[t]
\small
\centering
\resizebox{0.96\textwidth}{!}{
\begin{tabular}{l|ccc|ccc|c}
\toprule
{\multirow{2}{*}{\textbf{\makecell[l]{\\System}}}}
& \multicolumn{3}{c|}{\textbf{CoNLL14}}
& \multicolumn{3}{c|}{\textbf{BEA19 (test)}} 
& \multicolumn{1}{c}{\textbf{JFLEG (test)}} \\
\cmidrule(r){2-8}
& {Pre.}  & {Rec.}  & \textsc{F$_{0.5}$}   & {Pre.}  & {Rec.}  & \textsc{F$_{0.5}$}  & {GLEU} \\
\cmidrule(lr){1-8}
Transformer &60.1 &36.6 &53.3   &60.9 &48.3 &57.9   & 55.4 \\
TagGEC \citep{stahlberg-kumar-2021-synthetic}     &72.8 &49.5 &66.6   &72.1 &64.4 &70.4   & \bf 64.7 \\
GECToR      &\bf 75.6 &44.5 &66.3   &\bf 76.7& 57.8 &71.9   & 58.6  \\ 
T5 large    &72.2 &51.4 &66.8   &73.4 &67.0 &72.0   & 62.8 \\ 
T5 xxl \citep{rothe-etal-2021-simple}  &- &- &\bf 68.9   &- &- &\bf 75.9   & - \\
\cmidrule(lr){1-8}

ChatGPT (zero-shot CoT) &50.2 &59.0 &51.7    &32.1 &\bf 70.5* &36.1   & 61.4   \\
ChatGPT (1-shot CoT) &52.0 &58.1 &53.1    &34.6 &69.7 &38.4*   & 59.7  \\
ChatGPT (3-shot CoT) &51.3 &\bf 62.4* &53.2*    &34.0 &70.2 &37.9   & 63.5* \\
ChatGPT (5-shot CoT) &50.9 &61.8 &52.8    &32.4 &69.9 &36.3   & 62.5 \\

\bottomrule
\end{tabular}} 
\linespread{1}
\caption{Zero-shot CoT and few-shot CoT evaluation results with ChatGPT on the English CoNLL14, BEA19 test, and JFLEG test sets. 
The comparison GEC models for Transformer, GECToR and T5-large are ones that we trained on the English CLang8 data as \citet{rothe-etal-2021-simple}. \textbf{Bold} values indicate the best scores across different systems. * denotes the best results among different shot paradigms for ChatGPT.
}
\label{Tab:GEC_English}
\end{table*}


\begin{table*}[t]
\small
\centering
\resizebox{0.85\textwidth}{!}{
\begin{tabular}{l|ccc|ccc}
\toprule
{\multirow{2}{*}{\textbf{\makecell[l]{\\System}}}}
& \multicolumn{3}{c|}{\textbf{De (Falko-MERLIN)}}
& \multicolumn{3}{c}{\textbf{Zh (NPLCC18)}} \\
\cmidrule(r){2-7}
& {Pre.}  & {Rec.}  & \textsc{F$_{0.5}$}   & {Pre.}  & {Rec.}  & \textsc{F$_{0.5}$}  \\
\cmidrule(lr){1-7}
Transformer & 58.8 & 34.3 & 51.5 & 31.2 & 20.2 & 28.1 \\
GECToR  & - & - & - & 37.4 & 26.3 & 34.5 \\ 
mT5 large & \bf 75.4 & 55.1 & 70.2 & \bf 41.5 & 25.8 &\bf 37.0 \\ 
mT5 xxl  \citep{rothe-etal-2021-simple} & - & - & 74.8 & - & - & - \\ 
gT5 xxl  \citep{rothe-etal-2021-simple} & - & - & \bf 76.0 & - & - & - \\ 

\cmidrule(lr){1-7}

ChatGPT (zero-shot CoT) & 59.9 & 63.9 & 60.7  & 28.7 & 39.4 & 28.7* \\
ChatGPT (1-shot CoT) & 61.6 & 65.6 & 62.4   & 25.7 & \bf 40.8* & 27.8 \\
ChatGPT (3-shot CoT) & 63.1 & 65.3 & 63.5*   & 26.4 & 39.4 & 28.3 \\
ChatGPT (5-shot CoT) & 61.5 & \bf 65.9* & 62.3   & 25.2 & 39.2 & 27.2 \\

\bottomrule
\end{tabular}} 
\linespread{1}
\caption{Zero-shot CoT and few-shot CoT evaluation results with ChatGPT on the German Falko-MERLIN and Chinese NLPCC18 test sets. 
The comparison GEC models for Transformer, GECToR and T5-large are ones that we trained on the German CLang8 and Chinese lang8 datasets. \textbf{Bold} values indicate the best scores across different systems. * denotes the best results among different shot paradigms for ChatGPT.
}
\label{Tab:GEC_German_Chinese}
\end{table*}
		

We evaluate the English GEC task by conducting experiments on the CoNLL4, BAE19 test, and JFLEG test. 
Our experiment covers various shot paradigms, including zero-shot CoT, 1-shot CoT, 3-shot CoT, and 5-shot CoT. 

Tabel~\ref{Tab:GEC_English} shows the overall performance of ChatGPT with these shot paradigms. 
We can obtain some interesting observations from the results. 
Firstly, although there is a large margin between ChatGPT and existing GEC SOAT systems in terms of \textbf{F$_{0.5}$} and \textbf{Precision} scores on the CoNLL14 and BEA19 test sets, ChatGPT surpasses all the previous SOTA systems in \textbf{Recall} scores. 
This indicates that ChatGPT has a higher propensity for error detection and self-correction.

Moreover, ChatGPT performs extremely well on the JFLEG test set. 
Under the 3-shot CoT strategy, its \textbf{GLEU} score is very close to the SOTA score (falling only 1.2 points short) and it even surpasses the T5 large GEC system by 0.7 points. 
As we know, JFLEG is a dataset based on fluency editing, which can be used to evaluate a GEC system's ability to correct grammatical errors in a sentence while maintaining fluency. 
The promising scores obtained by ChatGPT demonstrate its potential for achieving fluency and naturalness in English sentence correction.

Additionally, comparing the performance of different shot CoT strategies on ChatGPT, the few-shot CoT prompting method outperforms the zero-shot CoT prompting method significantly, except for the 1-shot CoT prompting method on the JFLEG test set. 
This further demonstrates the effectiveness of the in-context learning method. 
Interestingly, providing more in-context examples does not necessarily lead to better performance for ChatGPT. 
Our experiments show that performance tends to decrease when the number of in-context examples exceeds five.

\subsection{ChatGPT Performance on non-English and Low-Resource GEC Tasks}

We also evaluate non-English GEC tasks on the ChatGPT system by conducting experiments on the German Falko-MERLIN test and Chinese NLPCC18 test sets. 
It is worth noting that the German training dataset is much smaller than that of English and Chinese, which means that GEC in the German language can be considered a low-resource task. 
Table~\ref{Tab:GEC_German_Chinese} presents the evaluation results of zero-shot CoT and few-shot CoT prompting methods. 
The results of German and Chinese GEC exhibit similar trends to those in English GEC, where ChatGPT outperforms the SOTA systems in \textbf{Recall} scores but scores significantly lower in \textbf{F$_{0.5}$} and \textbf{Precision}. 

Furthermore, ChatGPT surpasses the performance of the Transformer base model trained from scratch on the evaluation metrics, suggesting that with proper prompting methods, ChatGPT can effectively perform GEC on low-resource and non-English tasks. 
This also highlights the potential of ChatGPT for application in multilingual GEC tasks. 
Another interesting finding is that the performance of zero-shot CoT and few-shot CoT strategies in Chinese is opposite to that in English and German. 
Specifically, the few-shot CoT prompting method performs comparatively worse than the zero-shot CoT prompting method in Chinese.
We hypothesize that, on the one hand, ChatGPT is an LLM centered around English, and German and English belong to the same language family. 
Therefore, ChatGPT performs similarly on GEC tasks in both languages.
On the other hand, the lexicon of Chinese GEC is much more complex than that of English. 
However, this does not mean that ChatGPT's performance on Chinese GEC is limited to this extent. 
It may be necessary to design more effective few-shot selection methods to guide ChatGPT and improve its performance.

\begin{table*}[t]
\small
\centering
\resizebox{0.96\textwidth}{!}{
\begin{tabular}{l|ccc|ccc|ccc}
\toprule
{\multirow{2}{*}{\textbf{\makecell[l]{\\System}}}}
& \multicolumn{3}{c|}{\textbf{CoNLL14}} 
& \multicolumn{3}{c|}{\textbf{FCE (test)}}
& \multicolumn{3}{c}{\textbf{BEA19 (dev)}} \\
\cmidrule(r){2-10}
& {Pre.}  & {Rec.}  & \textsc{F$_{0.5}$}   & {Pre.}  & {Rec.}  & \textsc{F$_{0.5}$}  & {Pre.}  & {Rec.}  & \textsc{F$_{0.5}$} \\
\cmidrule(lr){1-10}
SingEnc & 59.8 & 27.3 & 48.3 & 61.6 & 45.0 & 57.4 & 57.0 & 43.2 & 53.5 \\ 
MultiEnc-enc & 63.2 & 28.0 & 50.5 & 65.6 & 42.7 & 59.2 & 62.1 & 41.7 & 56.5 \\ 
MultiEnc-dec & \bf 64.6 & 28.7 & \bf 51.6 & \bf 65.4 & 44.2 & \bf 59.7 &\bf 62.6 & 40.7 & \bf 56.6 \\ 
\cmidrule(lr){1-10}

ChatGPT(zero-shot CoT) & 42.3 & \bf 40.8 & 42.0 & 46.3 &\bf 49.9 & 47.0 & 47.7 & \bf 50.8 &48.3 \\
\bottomrule
\end{tabular}} 
\linespread{1}
\caption{Zero-shot CoT evaluation results with ChatGPT on the English CoNLL14 document-level test, FCE document-level test, and BEA19 document-level development sets. The comparison document-level GEC results for SingEnc, MultiEnc-enc, and MultiEnc-dec models are reported by \citet{yuan-bryant-2021-document}, where MultiEnc-dec is the current SOTA model.
\textbf{Bold} values indicate the best scores across different systems.
}
\label{Tab:doc_level_GEC}
\end{table*}
		

\begin{table}[t]
\small
\centering
\resizebox{0.45\textwidth}{!}{
\begin{tabular}{l|ccc}
\toprule
{\multirow{2}{*}{\textbf{\makecell[l]{\\Method}}}}
& \multicolumn{3}{c}{\textbf{BEA19 (dev)}} \\
\cmidrule(r){2-4}
 & {Pre.}  & {Rec.}  & \textsc{F$_{0.5}$}    \\

\cmidrule(lr){1-4}

ChatGPT (zero-shot CoT) & \bf 47.7 &  50.8 & \bf 48.3$^{\dagger}$ \\
ChatGPT (1-shot CoT) & 42.5 & \bf 55.6 &  44.6 \\

\bottomrule
\end{tabular}} 
\linespread{1}
\caption{Performance of zero-shot CoT and 1-shot CoT prompting methods for ChatGPT on
the BEA19 document-level development set. \textbf{Bold} values indicate the best scores. Statistically significant improvement over the 1-shot CoT prompting method is reported using $P${\_}value, $^{\dagger}p<0.01$.
}
\label{Tab:doc_GEC_1_shot}
\end{table}
		

\subsection{Document-Level GEC Performance with ChatGPT}

Prior research has shown limited attention toward document-level GEC task, with most studies concentrating on sentence-level GEC tasks using PLMs. 
As far as we know, only two studies have explored the improvement of performance in English document-level GEC \citep{chollampatt-etal-2019-cross, yuan-bryant-2021-document}, which only trained on CNN/Transformer architecture from scratch with their approach. 
In this section, we evaluate English document-level GEC tasks on the ChatGPT system by conducting experiments on the CoNLL14 document-level test, FCE document-level test, and BEA19 document-level development sets followed by \citet{yuan-bryant-2021-document}. 
Regarding the design of prompts, we simply replace ``\texttt{sentence}'' with ``\texttt{document}'' in Table~\ref{Tab:GEC_zero_shot_prompts} and Figure~\ref{fig:few-shot}.

Table~\ref{Tab:doc_level_GEC} presents the evaluation results of the zero-shot CoT prompting method for ChatGPT. 
Similar to sentence-level GEC tasks, ChatGPT exhibits good performance in terms of \textbf{Recall}, but its scores in \textbf{F$_{0.5}$} and \textbf{Precision} are significantly lower in the context of document-level GEC models that are trained from scratch on Transformer without leveraging knowledge from large language models. 
It is worth noting that this poor performance raises concerns and warrants further investigation.

Furthermore, the large number of sentences in each document has impeded our exploration of the few-shot CoT prompting method. 
To address this issue, we only conduct 1-shot CoT experiments using the BEA19 development set, which has the fewest average number of sentences compared to the other two test sets. 
For in-context examples, we randomly select one sample from the FCE test set. 
As shown in Table~\ref{Tab:doc_GEC_1_shot}, it appears that the 1-shot CoT strategy is not effective for document-level GEC. 
While there is a noticeable increase in \textbf{Recall}, the \textbf{F$_{0.5}$} and \textbf{Precision} scores still decreased significantly. 
In light of these results, we speculate that ChatGPT may not be sufficiently capable of processing long sentences, which often require high levels of coherence and consistency between sentences. 
In the subsequent analysis section~\ref{sec:doc-analysis}, we will examine ChatGPT's shortcomings from the perspective of correcting different types of errors.

\section{Human Evaluation and Analysis}\label{human}

\begin{table}[!t]
\small
\centering
\resizebox{0.45\textwidth}{!}{
\begin{tabular}{l|c|c}
\toprule
{\multirow{2}{*}{\textbf{\makecell[l]{\\System}}}}
& \multicolumn{1}{c|}{\textbf{CoNLL14}}
& \multicolumn{1}{c}{\textbf{JFLEG}} \\
\cmidrule(r){2-3}
& \textsc{F$_{0.5}$} & GLEU  \\
\cmidrule(lr){1-3}
Human & 72.58   & 62.37\\
\cmidrule(lr){1-3}
Transformer & 66.97  & 55.41 \\
GECToR  & 80.49  & 58.61 \\ 
T5 large & \bf 81.19 & 62.82 \\ 

\cmidrule(lr){1-3}

ChatGPT (0-shot) & 69.74 & 61.42\\
ChatGPT (1-shot) & 71.55 & 59.65 \\
ChatGPT (3-shot) & 71.73* &\bf 63.52* \\
ChatGPT (5-shot) & 70.66 & 62.53\\

\bottomrule
\end{tabular}} 
\linespread{1}
\caption{Zero-shot CoT and few-shot CoT Prompting methods for ChatGPT in comparison to automatic human evaluation performance.
\textbf{Bold} values indicate the best scores across different systems. * denotes the best results among different shot paradigms for ChatGPT.
}
\label{Tab:sysVShuamn}
\end{table}
		

\subsection{Automatic Human Evaluation}

\begin{figure*}[ht]
\centering
\includegraphics[width=1.0\textwidth, trim=0 0 0 0]{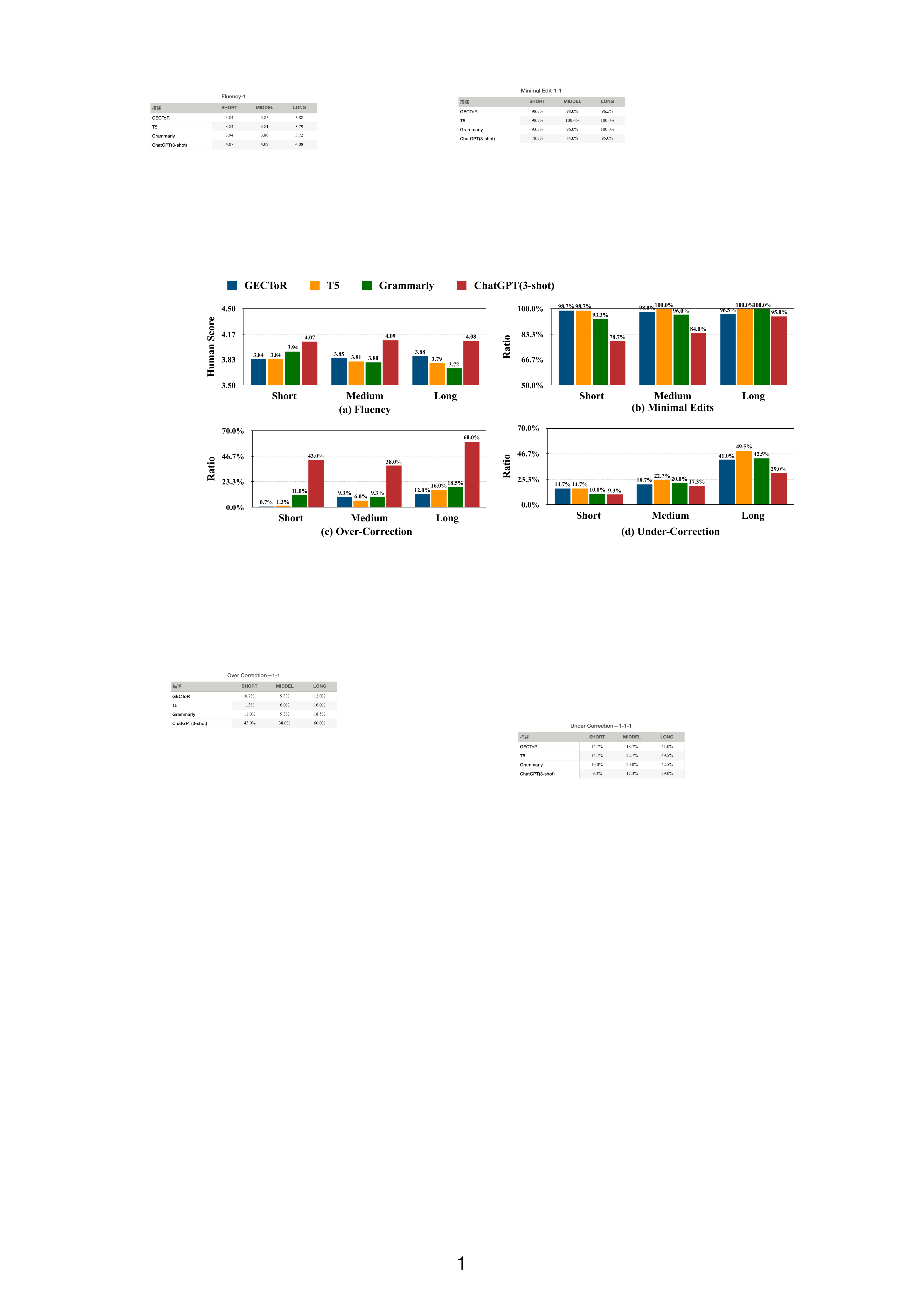}
\caption{The results of manual human evaluations on four criteria: Fluency, Minimal Edits, Over-Correction, and Under-Correction. The statistical analysis is based on the average scores provided by three evaluators.}
\label{fig:human_eval}
\end{figure*}

\citet{bryant-ng-2015-far} is the first to attempt to measure human performance on the CoNLL14 test set using 10 references. 
Specifically, they calculate the performance of each annotator by comparing its corrections to the other 9 annotators. The average of the 9 F$_{0.5}$ scores was taken as the final score for the human-level performance. 
To perform an automatic human evaluation on the ChatGPT performance, we adopt the approach used by \citet{bryant-ng-2015-far} and \citet{napoles-etal-2017-jfleg} to compare the performance of different GEC systems with that of humans on the CoNLL14 and JFLEG test sets. 
The reported F$_{0.5}$ scores for human performance on these test sets are 72.58 and 62.37, respectively.

Table~\ref{Tab:sysVShuamn} shows the automatic human evaluation results. 
We have observed some interesting phenomena. 
Firstly, the F$_{0.5}$ score of ChatGPT on the CoNLL14 test is lower than that of the Transformer base model in Table~\ref{Tab:GEC_English}, but the human evaluation shows a significant improvement for ChatGPT. 
This suggests that the existing evaluation methods may underestimate the performance of ChatGPT on sentence-level GEC tasks. 
Although the best few-shot CoT strategy for ChatGPT achieved a human evaluation score far lower than the scores of two mainstream SOTA models, it is only 0.85 F$_{0.5}$ points lower than the human-level evaluation score, indicating that ChatGPT has great potential in the sentence-level GEC task. 

Furthermore, the human evaluation performance of ChatGPT on the JFLEG test set is impressive. 
The best few-shot CoT prompting method not only exceeds the human-level evaluation score by 1.15 GLEU points but also outperforms the strong baseline T5 large model by 0.7 GLEU points. 
This observation suggests that the sentences corrected by ChatGPT exhibit a high level of fluency and naturalness.

\subsection{Manual Human Evaluation}
To explore ChatGPT's true performance on the CoNLL14 test set, we carry out a manual human evaluation.  
In addition to comparing with two mainstream SOTA GEC systems (GECToR, T5 large), we also consider the widely-used commercial system, Grammarly\footnote{\url{https://app.grammarly.com}}, renowned for its ability to detect and correct a range of errors in English texts, such as spelling, punctuation, grammar, and word choice, using features like  Correctness, Clarity, Engagement, and Delivery. 
In this evaluation, we only use its free and open Correctness feature to correct sentences.
We select the 50 longest, 50 medium-length, and 50 shortest sentences from the CoNLL14 test set and invite three postgraduate students with international study experience to conduct a manual evaluation. 
Our reference is the widely-used \emph{official-2014.combined.m2} version \citep{ng-etal-2014-conll} of CoNLL14, and we obtain the final evaluation by averaging the scores from the three students. 
The evaluation focus on four main categories: Fluency, Minimal Edits, Over-Correction, and Under-Correction. Some examples are shown in Appendix~\ref{appendix:examples}.

\paragraph{Fluency} refers to ensuring that the corrected sentence conforms to the linguistic conventions, is easily comprehensible, flows naturally, and preserves the intended meaning of the original sentence while correcting the grammatical errors. We define a 1-5 rating scale, where 1 represents the lowest and 5 is the highest level of fluency. From the results shown in Figure~\ref{fig:human_eval}(a), it can be seen that ChatGPT exhibits significantly better fluency in correcting grammar errors of short, medium, and long sentences compared to the other three mainstream systems. 
In addition, we also observed that among the three mainstream systems, Grammarly performs more fluently in correcting grammar errors in short sentences, while GECToR performs better in long sentences. As for sentences of medium length, the three systems perform similarly. 

\paragraph{Minimal Edits}
The CoNLL14 test set includes grammatically incorrect sentences that have been created by making minimal edits to grammatically correct sentences. 
We compare the output of ChatGPT and three other systems to the references sentences in order to determine whether they use minimal edits. 
The annotators assign a value of 1 to indicate that a sentence followed minimal edits, and 0 to indicate that it does not.  
We compute the proportion of minimal edits for sentences of different lengths. The results are shown in Figure~\ref{fig:human_eval}(b). 
Compared to the other three systems, ChatGPT is not particularly inclined to use minimal edits, especially in short sentences. 
However, as the length of the sentence increases, ChatGPT becomes more willing to follow minimal edits.
Furthermore, Grammarly exhibits a lower level of conformity to minimal edits in short sentences compared to two mainstream SOTA systems. 
In terms of medium and long sentences, GECToR shows a lower level of minimal edits compared to T5 large.
Interestingly, comparing Figure~\ref{fig:human_eval}(a), we find that the less a system adheres to minimal edits, the better the fluency performance of the generated sentences.

\paragraph{Over-Correction}
Since there can be multiple ways to correct a sentence, sometimes a correction that differs from the references does not necessarily mean the sentence has not been corrected. 
Over-Correction is used to evaluate a system's ability to produce correct results beyond what is indicated by the reference correction. 
We also ask the annotators to assign a value of 1 to indicate that a sentence is an over-correction, and 0 to indicate that it is not. 
We calculate the proportion of Over-Correction for sentences with varying lengths and the results are depicted in Figure~\ref{fig:human_eval}(c). 
ChatGPT surpasses the other three GEC systems in its capacity to over-correct sentences of different lengths, which indicates that it can correct sentences freely and diversely, which is consistent with its good fluency demonstrated in Figure~\ref{fig:human_eval}(a). 
Additionally, we observe that Grammarly also exhibits over-correction compared to GECToR and T5 large models.

\paragraph{Under-Correction} refers to cases where a GEC system fails to correct certain grammatical errors that exist in the original sentence. Our requirement for annotators is to check any errors present in the generated sentences by GEC systems that are not identified or corrected. 
Similar to Over-Correction, 1 indicates a sentence is an under-correction, and 0 indicates that it is not. Interestingly, in contrast to Over-Correction, ChatGPT demonstrates fewer Under-Corrections than the other systems in short, medium, and long sentences, as shown in Figure~\ref{fig:human_eval}(d). 
This suggests that ChatGPT has great potential for GEC tasks. 
Furthermore, T5 large GEC system is prone to Under-Corrections, which highlights that a higher F$_{0.5}$ score does not imply a more perfect GEC system. 
We also observe that as sentence length increases, ChatGPT exhibits a stronger tendency towards Under-Correction. 
This may explain its poor performance in document-level GEC tasks.

Through manual analysis, we can conclude that ChatGPT generates grammatically corrected sentences that are highly fluent, which may be attributed to its powerful diversity and free-generation ability.

\begin{table*}[t]
\small
\centering
\resizebox{0.98\textwidth}{!}{
\begin{tabular}{l|ccc|ccc|ccc|r}
\toprule
{\multirow{2}{*}{\textbf{\makecell[l]{\\Error-Type}}}}
& \multicolumn{3}{c|}{\textbf{MultiEnc-dec}} 
& \multicolumn{3}{c|}{\textbf{ChatGPT(zero-shot)}}
& \multicolumn{3}{c|}{\textbf{ChatGPT(1-shot)}} 
& \multicolumn{1}{c}{\textbf{Differ.}} \\
\cmidrule(r){2-11}
& {Pre.}  & {Rec.}  & \textsc{F$_{0.5}$}   & {Pre.}  & {Rec.}  & \textsc{F$_{0.5}$}  & {Pre.}  & {Rec.}  & \textsc{F$_{0.5}$} & \textsc{F$_{0.5}$}\\
\cmidrule(lr){1-11} 
ADJ      & 44.4 & 15.6 & \bf 32.5    &34.9 &20.2 &30.4   &32.2 &32.5 &32.3   &-0.2 \\
ADJ:FORM & 100.0 & 25.0 & 62.5   &60.0 &75.0 &62.5   &56.5 &81.3 &60.2   & 0.0 \\
ADV      & 42.1 & 17.9 & \bf 33.2    &18.7 &17.7 &18.5   &18.6 &27.7 &19.9   &-13.3 \\
CONJ     & 46.2 & 14.3 & \bf 31.9    &13.2 &22.5 &14.4   &15.0 &37.5 &17.1   &-14.8 \\
CONTR    & 85.0 & 58.6 & \bf 78.0    &38.7 &38.7 &38.7   &28.6 &51.6 &31.4   &-39.3 \\
DET      & 63.7 & 51.3 & \bf 60.8    &57.9 &59.1 &58.1   &54.1 &66.0 &56.1   &-2.7 \\
MORPH    & 70.6 & 33.3 & \bf 57.7    &44.6 &55.6 &46.5   &45.2 &65.6 &48.2   &-9.5 \\
NOUN     & 38.1 & 13.3 & 27.7    &41.2 &21.5 &\bf 34.8   &31.5 &26.9  &30.4  &\bf +7.1 \\
NOUN:INFL & 100.0 & 75.0 & \bf 93.8  &88.9 &72.7 &85.1   &90.9 &90.9 &90.9   &-2.9 \\
NOUN:NUM  & 74.2 & 47.6 & \bf 66.7   &59.1 &60.2 &59.3   &58.1 &74.5 &60.7   &-6.0 \\
NOUN:POSS & 63.0 & 51.8 & 60.4   &68.4 &65.0 &67.7   &69.0 &66.7 & \bf 68.5  &\bf +0.8 \\
ORTH      & 71.2 & 59.9 & \bf 68.6   &64.3 &76.1 &66.4   &63.3 &78.6 &65.8   &-2.2 \\
OTHER     & 38.1 & 22.5 & \bf 33.5   &26.5 &29.8 &27.1   &18.8 &33.5 &20.6   &-6.4 \\
PART      & 58.5 & 40.7 & 53.8   &71.4 &39.7 & \bf 61.6   &60.3 &55.6 &59.3  &-7.8 \\
PREP      & 64.7 & 41.9 & \bf 58.3   &56.6 &43.2 &53.3   &49.4 &52.1 &49.9   &-5.0 \\
PRON      & 55.7 & 43.1 & \bf 52.6   &43.1 &53.9 &44.9   &33.6 &56.9 &36.6   &-7.7 \\
PUNCT     & 70.1 & 47.3 & \bf 63.9   &42.2 &63.2 &45.2   &42.8 &62.2 &45.6   &-18.3 \\
SPELL     & 86.2 & 53.9 & 76.9   &81.5 &89.3 & \bf 82.9   &81.3 &88.6 &82.7  &\bf +6.0 \\
VERB      & 44.8 & 21.6 & \bf 36.9   &39.5 &21.9 &34.0   &31.5 &31.2 &31.5   &-2.9 \\
VERB:FORM & 71.1 & 60.9 & \bf 68.8   &50.5 &68.6 &53.3   &45.5 &68.6 &48.8   &-15.5 \\
VERB:INFL & 80.0 & 66.7 & 76.9   &80.0 &66.7 & 76.9   &57.1 &66.7 &58.8   &0.0 \\
VERB:SVA  & 72.4 & 74.5 & \bf 72.8   &64.7 &82.9 &67.7   &61.8 &84.3 &65.3   &-5.1 \\
VERB:TENSE & 63.5 & 44.8 & \bf 58.6  &58.1 &44.5 &54.8   &55.4 &53.2 &54.9   &-3.7 \\
WO         & 64.7 & 37.5 & \bf 56.5  &41.0 &46.6 &42.0   &36.2 &47.7 &38.0   &-14.5 \\
\cmidrule(lr){1-11}
Total      & 62.6 & 40.7 & \bf 56.6  &47.7 &50.8 &48.3   &42.5 &55.6 &44.63  &-8.3 \\

\bottomrule
\end{tabular}} 
\linespread{1}
\caption{Fine-grained error types performance of \textbf{ChatGPT} and \textbf{MultiEnc-dec} systems on the BEA19 development set. \textbf{Differ.} refers to the difference in F$_{0.5}$ scores between the two systems.
\textbf{Bold} values indicate the best F$_{0.5}$ scores.
}
\label{Tab:doc_analysis}
\end{table*}
		

\subsection{Fine-grained Error Analysis for Document-Level GEC}\label{sec:doc-analysis}
To evaluate ChatGPT's ability to correct various types of errors in document-level GEC, we conduct an study based on \citet{yuan-bryant-2021-document}, using the ERRANT toolkit \citep{bryant-etal-2017-automatic} to analyze the results on the BEA19 document-level development set. The results are presented in Table~\ref{Tab:doc_analysis}, comparing the performance of ChatGPT and the MultiEnc-dec document-level GEC system on POS-based fine-grained error types. 
ChatGPT demonstrates strong performance in addressing errors related to punctuation, nouns, and possessive nouns, achieving improvements of \textbf{6.0}, \textbf{7.1}, and \textbf{0.8 F$_{0.5}$} scores, respectively.
Regarding errors that frequently demand a strong understanding of agreement, coreference, or tense across multiple sentences, these are typically considered document-level errors. 
The ChatGPT performs significantly worse than the MultiEnc-dec model in correcting errors related to subject-verb agreement, prepositions, noun numbers, determiners, and pronouns (e.g. \texttt{VERB:SVA} \textbf{-5.1 F$_{0.5}$}, \texttt{VERB:TENSE} \textbf{-3.7 F$_{0.5}$}, \texttt{VERB:FORM} \textbf{-15.5 F$_{0.5}$}, \texttt{NOUN:NUM} \textbf{-6.0 F$_{0.5}$}, \texttt{PRON} \textbf{-7.7 F$_{0.5}$}.).  Moreover, ChatGPT's performance was even worse in handling cross-sentence boundary errors, as evidenced by a decline of \textbf{14.8 F$_{0.5}$} in \texttt{CONJ} and \textbf{18.3 F$_{0.5}$} in \texttt{PUNCT}.
We speculate the underperformance of the ChatGPT could be attributed to its potential limitations in terms of contextual memory, consistency, and coherence across sentences at the document-level.

\section{Conclusion}
In this study, we undertake a thorough examination of ChatGPT's performance and potential in GEC, as well as compare it with current SOTA models. To the best of our knowledge, we are the first to design zero-shot CoT and few-shot CoT settings for ChatGPT in the GEC task. Our experiments involve evaluating ChatGPT on five official test sets in three different languages, as well as on three document-level GEC test sets in English. The experimental results demonstrate that ChatGPT has strong error detection capabilities and can generate sentences with human-like fluency, despite its poor performance in precision and F$_{0.5}$ scores. Additionally, we find that ChatGPT also demonstrate its advantages in multilingual and low-resource settings, showing great potential for multilingual GEC. The results of further human evaluations once again confirm that ChatGPT is a highly fluent GEC system, and demonstrate that its performance can be better leveraged through the use of chain-of-thought prompts. However, our analysis of ChatGPT's ability to correct various types of errors in document-level GEC indicates that it performs poorly on most error types, such as agreement, coreference, tense errors across sentences, and cross-sentence boundary errors.

\section*{Limitations}

Our study has some limitations as follows: \\
\begin{itemize}
    \item   Our current work solely focus on exploring ChatGPT's potential performance in GEC, without delving into investigating other mainstream LLMs, such as other GPT-3.5 series models or the latest GPT-4 model from OpenAI. However, we plan to include these models in our future work.\\
    \item  Our study highlights that ChatGPT exhibits higher fluency and tends to make more casual edits in GEC tasks. However, in certain practical applications, such as language education, users may require only minimal editing corrections \citep{coyne2023analysis}. Therefore, exploring ChatGPT's potential in this aspect through the design of prompting methods is also a worthwhile pursuit.\\
   \item  For the few-shot settings using in-context learning, we follow \citet{vilar2022prompting} in randomly selecting examples from the development set, without specifically designing corresponding examples for each sentence. It is worth investigating whether selecting better examples could help improve ChatGPT's performance in GEC. \\
   \\
\end{itemize}

\section*{Acknowledgments}
This work was supported in part by the Science and Technology Development Fund, Macau SAR (Grant 
Nos. FDCT/0070/2022/AMJ, FDCT/060/2022/AFJ) and the Multi-year Research Grant from the University 
of Macau (Grant No. MYRG2020-00054-FST). This work was performed in part at SICC which is supported 
by SKL-IOTSC, and HPCC supported by ICTO of the University of Macau.

\bibliography{anthology,custom,custom_ref}
\bibliographystyle{acl_natbib}

\clearpage
\appendix

\section{Appendix}
\subsection{Examples of zero-shot settings}\label{appendix:zero-shot-settings}

\begin{figure*}[ht]
\centering
\includegraphics[width=0.9\textwidth, trim=0 0 0 0]{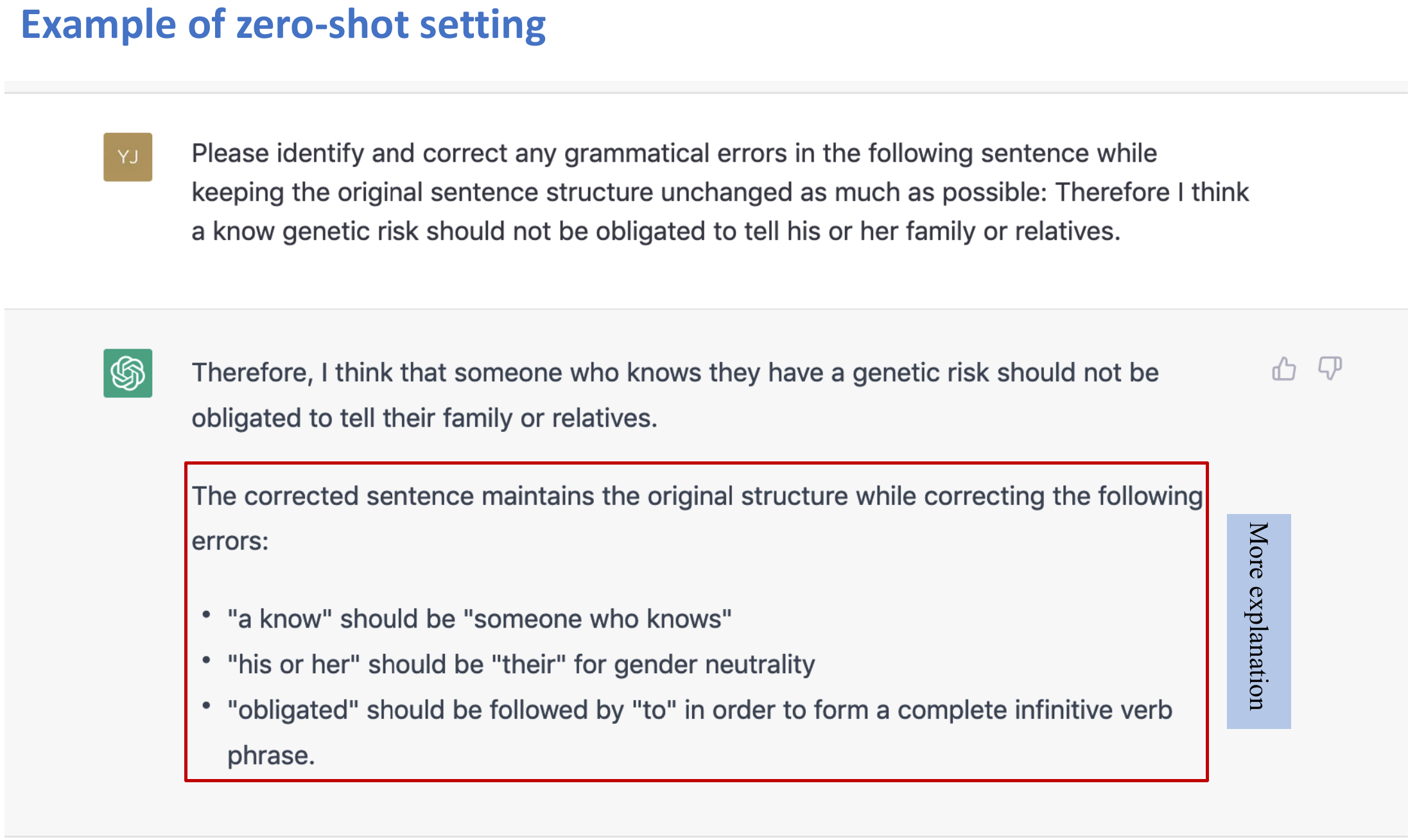}
\caption{Zero-shot setting for ChatGPT.
}
\label{fig:zero-shot-seting}
\end{figure*}

\begin{figure*}[ht]
\centering
\includegraphics[width=0.9\textwidth, trim=0 0 0 0]{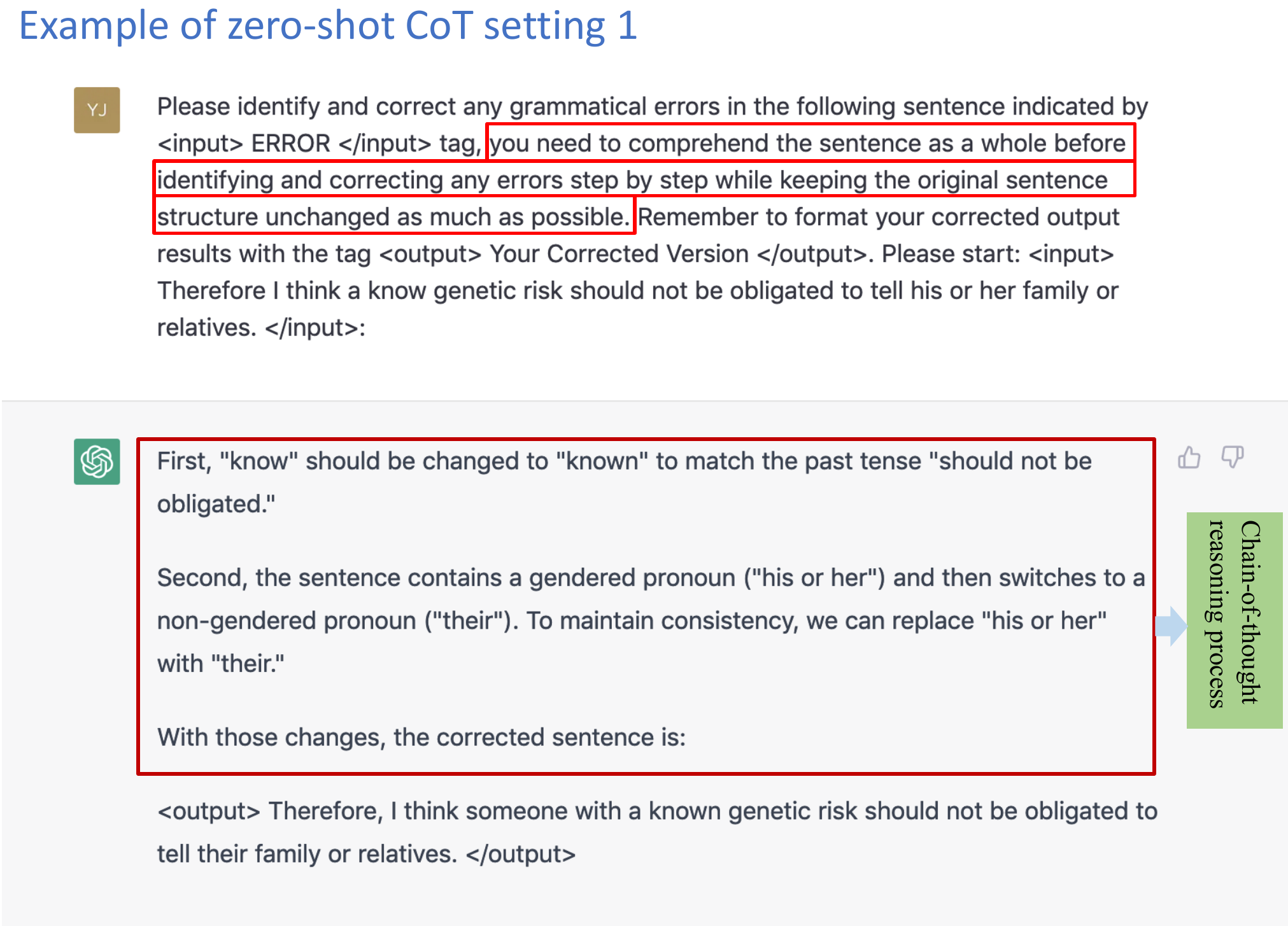}
\caption{Zero-shot CoT setting for ChatGPT. The prompts without uing ``afterward, output the corrected version directly without any explanations''.
}
\label{fig:zero-shot-cot-setting1}
\end{figure*}

\begin{figure*}[ht]
\centering
\includegraphics[width=0.9\textwidth, trim=0 0 0 0]{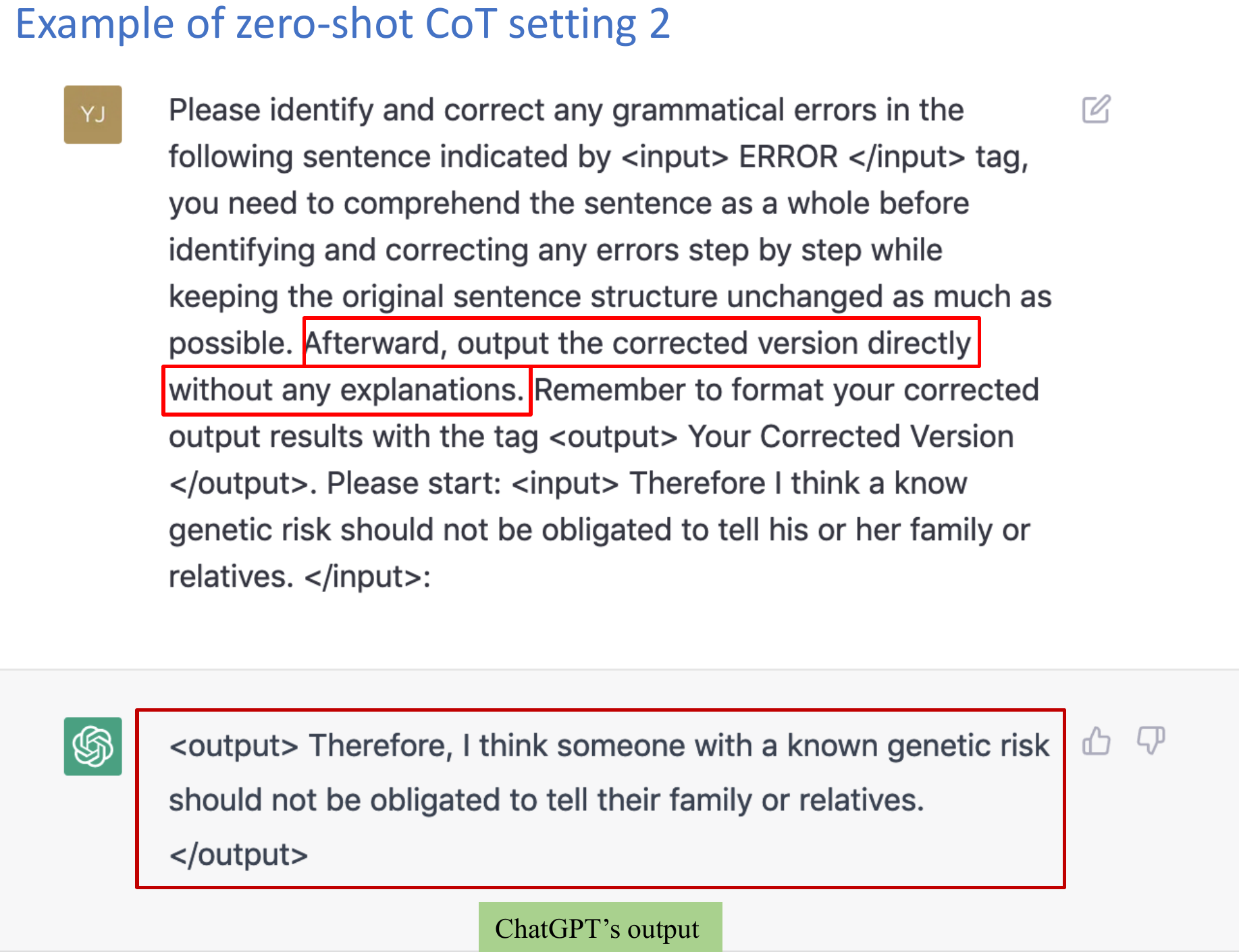}
\caption{Zero-shot CoT setting for ChatGPT.}
\label{fig:zero-shot-cot-setting2}
\end{figure*}

Figure~\ref{fig:zero-shot-seting} presents an example of the zero-shot setting for ChatGPT. Figure~\ref{fig:zero-shot-cot-setting1} and Figure~\ref{fig:zero-shot-cot-setting2} show the examples of zero-shot CoT settings for ChatGPT.

\subsection{GEC System Output Examples}\label{appendix:examples}

Table~\ref{Tab:case_study} presents some output examples generated by various system. It is noticeable that ChatGPT, in short and medium-length sentences, does not adhere strictly to the minimum editing principle and tends to over-correct. However, the resulting sentences exhibit higher fluency. In the case of long sentences, ChatGPT also demonstrates instances of over-correction, such as replacing "while" with "and."  While there is no grammatical error, the absence of a conspicuous contrasting relationship results in a decline in the overall sentence quality.

\begin{table*}[ht]
\small
\centering
\resizebox{.98\textwidth}{!}{
\begin{tabular}{p{0.5cm} | p{2.0cm}  p{8cm}  p{1.2cm} }
\toprule
%
 \bf Type & \bf System & \bf Examples & \bf Human \\
\cmidrule(r){1-4}

\multirow{6}{*}{{\textbf{S}}} & \multirow{1}{*}{{Src}} & It can make people become closer as well. &   \\
%
& \multirow{1}{*}{{Ref.}} & It can make people become closer as well. &   \\
%
%
%
& \multirow{1}{*}{{GECToR}} & It can make people become closer as well. & \multirow{1}{*}{4 | 1 | 0 | 0 }  \\
%
%
& \multirow{1}{*}{T5 large} & It can make people become closer as well. & \multirow{1}{*}{4 | 1 | 0 | 0 } \\
%
%
& \multirow{1}{*}{Grammarly} &It can make people become closer as well. & \multirow{1}{*}{4 | 1 | 0 | 0 }\\
\cdashline{2-4}
& \multirow{1}{*}{ChatGPT(3-shot)} &It can \textcolor{blue}{bring} people closer together as well. & \multirow{1}{*}{5 | 0 | 1 | 0 }\\
%
%
\cmidrule(r){1-4}
\multirow{12}{*}{{\textbf{M}}} & \multirow{2}{*}{{Src}} & Recent research has shown that the more people spend time on social media sites, the less they become ambitious. &   \\
%
& \multirow{2}{*}{{Ref.}} & Recent research has shown that the more people spend time on social media sites, the less they become ambitious. &   \\
%
%
%
& \multirow{2}{*}{{GECToR}} & Recent research has shown that the more people spend time on social media sites, the less they become ambitious. & \multirow{2}{*}{4 | 1 | 0 | 0 }  \\
%
%
& \multirow{2}{*}{T5 large} & Recent research has shown that the more people spend time on social media sites, the less they become ambitious. & \multirow{2}{*}{4 | 1 | 0 | 0 } \\
%
%
& \multirow{2}{*}{Grammarly} & Recent research has shown that the more people spend time on social media sites, the less they become ambitious. & \multirow{2}{*}{4 | 1 | 0 | 0 }\\
\cdashline{2-4}
& \multirow{2}{*}{ChatGPT(3-shot)} & Recent research has shown that the more time people spend on social media sites, \textcolor{blue}{the less ambitious they become}. & \multirow{2}{*}{5 | 0 | 1 | 0 }\\
%
%
%
\cmidrule(r){1-4}
\multirow{24}{*}{{\textbf{L}}} & \multirow{4}{*}{{Src}} & Though it is said that genetic testing involves emotional and social risks due to the test results, while the potential negative impacts of the risk still exist, the \textcolor{red}{consequence} will be significant if other members of his or her family do not know. &   \\
%
%
& \multirow{4}{*}{{Ref.}} & Though it is said that genetic testing involves emotional and social risks due to the test results, while the potential negative impacts of the risk still exist, the consequences will be significant if other members of his or her family do not know. &   \\
%
%
& \multirow{4}{*}{{GECToR}} & Though it is said that genetic testing involves emotional and social risks due to the test results, while the potential negative impacts of the risk still exist, the consequences will be significant if other members of his or her family do not know. & \multirow{4}{*}{4 | 1 | 0 | 0 }  \\
%
%
& \multirow{4}{*}{T5 large} & Though it is said that genetic testing involves emotional and social risks due to the test results, while the potential negative impacts of the risk still exist, the consequence will be significant if other members of his or her family do not know. & \multirow{4}{*}{4 | 1 | 0 | 0 } \\
%
%
& \multirow{4}{*}{Grammarly} & Though it is said that genetic testing involves emotional and social risks due to the test results, while the potential negative impacts of the risk still exist, the \textcolor{red}{consequence} will be significant if other members of his or her family do not know. & \multirow{4}{*}{4 | 1 | 0 | 1 }\\
\cdashline{2-4}
& \multirow{4}{*}{ChatGPT(3-shot)} & Though it is said that genetic testing involves emotional and social risks due to the test results, \textcolor{blue}{and} the potential negative impacts of the risk still exist, the consequences will be significant if other members of his or her family do not know. & \multirow{4}{*}{3 | 1 | 1 | 0 }\\
\bottomrule
\end{tabular}}
\linespread{1}
\caption{Different GEC systems output examples.  The words highlighted in \textcolor{red}{red} indicate grammatical errors, while those highlighted in \textcolor{blue}{blue} represent the corrected version by ChatGPT. \textbf{Human} denotes the evaluator's scores for Fluency | Minimal Edits | Over-Correction | Under-Correction. \textbf{S/ M/ L} stands for short/ medium/ long sentences.
}
\label{Tab:case_study}
\vskip -1em
\end{table*}


\end{document}